\documentclass[letterpaper, 10 pt, conference]{ieeeconf}  

\IEEEoverridecommandlockouts                              

\usepackage{multirow}
\usepackage{amsmath}  
\usepackage{amssymb}  
\usepackage{color,array}
\usepackage{xcolor}
\usepackage{graphicx}
\usepackage{setspace}  
\usepackage{makecell}
\usepackage{subcaption}
\usepackage{caption}
\DeclareCaptionLabelSeparator{periodspace}{.\quad}
\usepackage{graphics} 
\usepackage{epsfig} 
\usepackage{times} 
\usepackage{amsmath} 
\usepackage{amssymb}  
\usepackage[table]{xcolor}

\usepackage[linesnumbered,ruled,vlined]{algorithm2e}
\usepackage{subfloat}

\usepackage{booktabs} 
\usepackage{gensymb}  
\usepackage{bm}  
\usepackage{underscore}  

\usepackage{amsmath} 
\usepackage{amssymb}  

\usepackage{threeparttable}
\usepackage{enumerate}
\usepackage{bbm}

\usepackage[hidelinks]{hyperref}

\usepackage{bbding} 
\usepackage{cite}
\usepackage{hyperref}
\author{{Tianle Zeng, Jianwei Peng, Hanjing Ye, Guangcheng Chen, Senzi Luo, Hong Zhang, \textit{Life Fellow, IEEE} } 
\thanks{All authors are with the Shenzhen Key Laboratory of Robotics and Computer Vision, Southern University of Science and Technology.}%
\thanks{Corresponding author: Hong Zhang (hzhang@sustech.edu.cn).}
\thanks{Project Page: \url{https://tianlezeng.github.io/EzReal/}}%
}

\begin{document}
\title{\LARGE \bf
EZREAL: Enhancing Zero-Shot Outdoor Robot Navigation toward Distant Targets under Varying Visibility

}
\makeatletter
\let\@oldmaketitle\@maketitle%
\renewcommand{\@maketitle}{\@oldmaketitle%
    \centering
    \vspace*{1mm}
    \includegraphics[width=\textwidth]{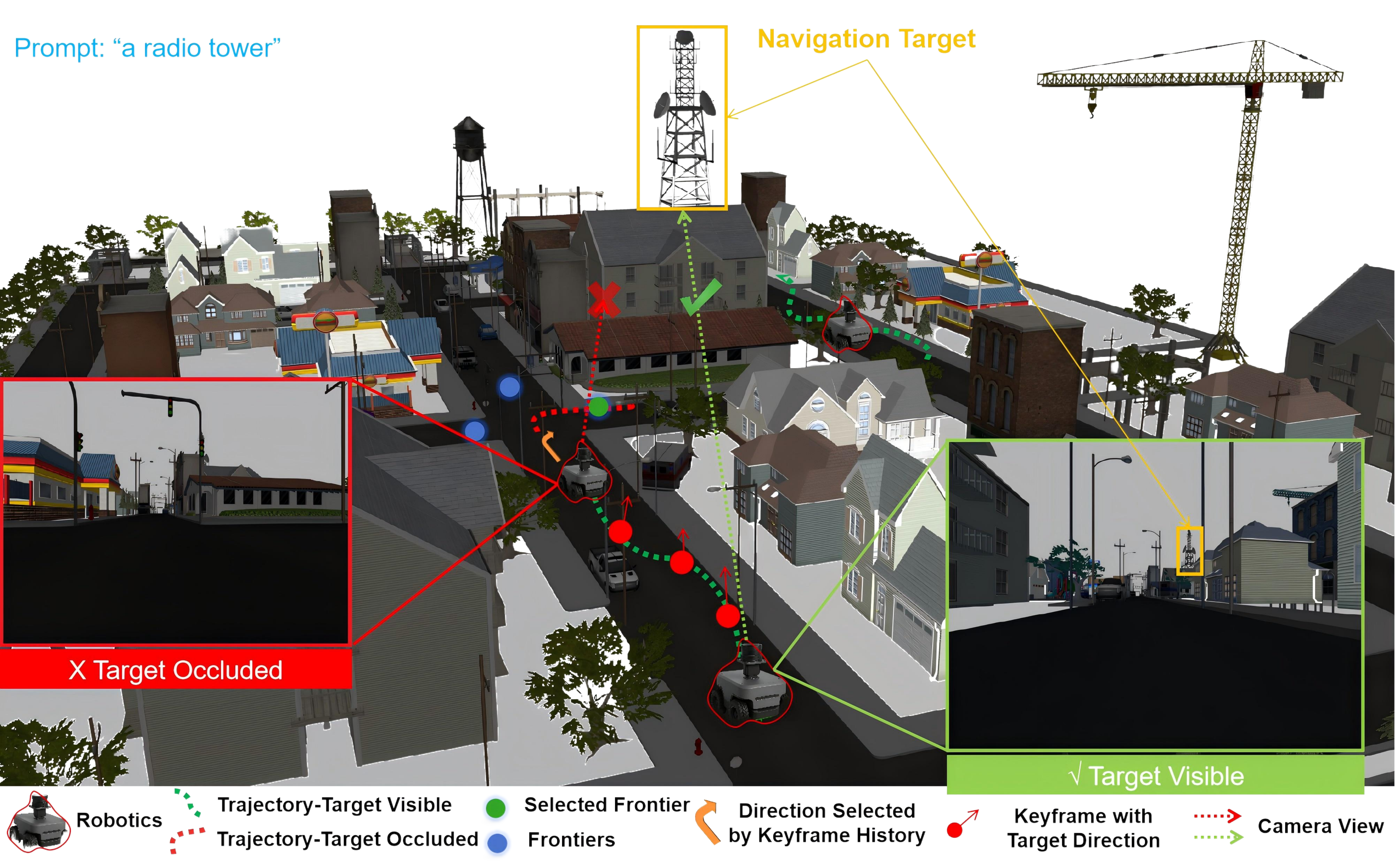}
    \captionof{figure}{Overview of our method for outdoor zero-shot object navigation. 
                The robot navigates toward a distant target specified by a language prompt. Our system achieves reliable semantic perception of tiny far-field targets and robust guidance under dynamic visibility conditions.}
    \label{whatwedo}
    \vspace*{-3mm}
}
\makeatother

\maketitle
\setcounter{figure}{1} 
\thispagestyle{empty}
\pagestyle{empty}

\begin{abstract}
Zero-shot object navigation (ZSON) in large-scale outdoor environments faces many challenges; we specifically address a coupled one: long-range targets that reduce to tiny projections and intermittent visibility due to partial or complete occlusion. We present a unified, lightweight closed-loop system built on an aligned multi-scale image tile hierarchy. Through hierarchical target–saliency fusion, it summarizes localized semantic contrast into a stable coarse-layer regional saliency that provides the target direction and indicates target visibility. This regional saliency supports visibility-aware heading maintenance through keyframe memory, saliency-weighted fusion of historical headings, and active search during temporary invisibility. The system avoids whole-image rescaling, enables deterministic bottom-up aggregation, supports zero-shot navigation, and runs efficiently on a mobile robot. Across simulation and real-world outdoor trials, the system detects semantic targets beyond 150\,m, maintains a correct heading through visibility changes with 82.6\% probability, and improves overall task success by 17.5\% compared with the SOTA methods, demonstrating robust ZSON toward distant and intermittently observable targets.
\end{abstract}


\section{INTRODUCTION}


Zero-shot object navigation (ZSON) enables robots to pursue open-vocabulary targets without category-specific training~\cite{zhang2024vision}. This ability is particularly valuable for high-impact applications such as exploration, search and rescue, field inspection, and post-disaster assessment~\cite{sun2024survey}. Existing ZSON methods~\cite{gadre2023cows,zhou2023esc,yokoyama2024vlfm,werby2024hierarchical} have shown strong performance in indoor environments, where scene structures are compact and semantic cues are dense. In contrast, transferring ZSON to large-scale outdoor environments remains highly challenging.


Unlike indoor ZSON, which typically relies on dense depth-based semantic maps as a navigation foundation, outdoor environments pose a fundamentally different challenge: target objects often lie far beyond the effective range of depth sensors, making depth-based semantic mapping infeasible. In such cases, robots must rely solely on RGB observations, where distant targets shrink to tiny projections with limited visual cues. This same geometry makes visibility inherently unstable: with only a few pixels, modest viewpoint changes or foreground structures can easily occlude the target, so visibility often fluctuates between faint, partial, and fully missing. Consequently, tiny far-field perception and dynamic visibility form a coupled failure mode, making it difficult to maintain a consistent heading from direct observation alone. Standard tiny-object detectors~\cite{muzammul2025comprehensive} do not address this navigation-specific coupling and are discussed in Section~\ref{relativework}.

While recent efforts~\cite{alama2025rayfronts,sridhar2024nomad,zheng2025gmm,shah2023gnm} have extended ZSON to outdoor settings through end-to-end training or Large Language Model reasoning-guided path planning, their approaches typically rely on implicit representations and do not explicitly address the coupled challenge of weak long-range cues under dynamic visibility changes. As a result, they often struggle with delayed decisions, degraded recognition of tiny distant targets, and unstable heading control under intermittent occlusion.

To tackle this unified challenge, we propose EZREAL, a lightweight ZSON system designed for long-range target navigation under varying visibility as illustrated in Fig.~\ref{whatwedo}. 
The system uses an aligned multi-scale image tile hierarchy with hierarchical target–saliency fusion to extract reliable directional cues from tiny, distant targets, and reuses the same regional saliency for visibility-aware heading maintenance even when targets become partially or fully occluded. 
By explicitly integrating these components in a unified closed loop, the system strengthens long-range semantic perception and ensures robust navigation heading maintenance across dynamic visibility conditions.

The main contributions of this work are summarized as follows:
\begin{itemize}
    \item We address the coupled challenge of long-range navigation under dynamic visibility changes, where targets appear as tiny projections and are frequently occluded. To this end, we design a lightweight zero-shot perception and navigation system that combines multi-scale semantic amplification with saliency-driven visibility detection and heading maintenance, ensuring robust guidance without category-specific training or fine-tuning.  

    \item We demonstrate the effectiveness of our system through integration into an occupancy-grid-based frontier exploration framework and extensive evaluation in both large-scale simulation and real-world environments, achieving significant gains in heading stability, occlusion recovery, and overall task success compared with the SOTA outdoor ZSON approaches.
\end{itemize}

\section{Related Work}
\label{relativework}
Zero-shot object navigation (ZSON) typically relies on semantic spatial representations to guide robots toward target objects. Common approaches~\cite{gadre2023cows,yokoyama2024vlfm,werby2024hierarchical,majumdar2022zson,zhou2023esc,wang2023gridmm} align depth and semantics to construct maps or scene graphs for localization and planning. Others~\cite{chen2022think,qiao2022hop,wang2023scaling,long2024discuss,zhou2024navgpt,chen2024mapgpt} leverage large-scale instruction datasets or large language models (LLMs) to reason over past observations. However, in large-scale outdoor settings, targets often lie beyond the depth sensing range, making strict depth–semantic alignment infeasible. Moreover, distant targets shrink to tiny projections with limited cues, making visibility unstable—often resulting in faint, partial, or full occlusion. Such semantic sparsity and intermittent visibility undermine both map-centric and LLM-driven pipelines. Taken together, outdoor ZSON presents a coupled challenge: far-field tiny-object perception and dynamic visibility must be handled jointly to maintain a consistent RGB-only heading.

Within computer vision, extremely small instances are studied as tiny object detection (TOD). Classical TOD improves recognition via multi-scale feature pyramids~\cite{lin2017feature,tan2020efficientdet}, label assignment~\cite{xu2022rfla}, and context modeling~\cite{hu2018relation,zhu2020deformable}, often in non-egocentric settings with dense targets and closed vocabularies, and with a per-frame focus~\cite{muzammul2025comprehensive}. TOD neither models intermittent visibility nor provides a stable world-frame heading when detections are sparse or missing. These assumptions diverge from outdoor ZSON, where a robot pursues a single, distant, open-vocabulary target under real-time egocentric constraints. Therefore, directly applying TOD is insufficient for outdoor ZSON, where perception must output target direction and visibility for real-time control.

Open-vocabulary vision experts~\cite{ren2024grounded,liu2024grounding,zhang2023faster,ranzinger2024radio} broaden category coverage but typically rely on global heatmaps and high-resolution cues, degrading sharply at few-pixel scales. Attention-enhanced or LLM-assisted variants~\cite{li2025dyfo, wu2024v} can sharpen focus and enable zero-shot detection, but their multi-stage pipelines are computationally heavy and impractical for real-time, on-board egocentric perception.

For outdoor-oriented ZSON, recent advances~\cite{alama2025rayfronts} explicitly project semantics beyond the observed volume to guide exploration, but uniform treatment of features across occlusion states can misinterpret cues and destabilize long-range guidance. Other methods~\cite{10687470,zheng2025gmm} leverage LLMs with structured prompts; under unstable visibility they often pause to reason whenever the target disappears—even during short occlusions—resulting in discontinuous navigation. Combined with prompt latency, these approaches struggle to support real-time deployment. More broadly, existing outdoor ZSON systems seldom model dynamic visibility explicitly, leaving guidance fragile under frequent occlusions.

In contrast, our system combines multi-scale semantic amplification with a saliency-driven heading memory, providing a lightweight and robust perception–action coupling that explicitly addresses the unified challenge of long-range outdoor ZSON under dynamic visibility.

\section{Method}
\begin{figure*}[!t]
    \centering
    \includegraphics[width=\textwidth]{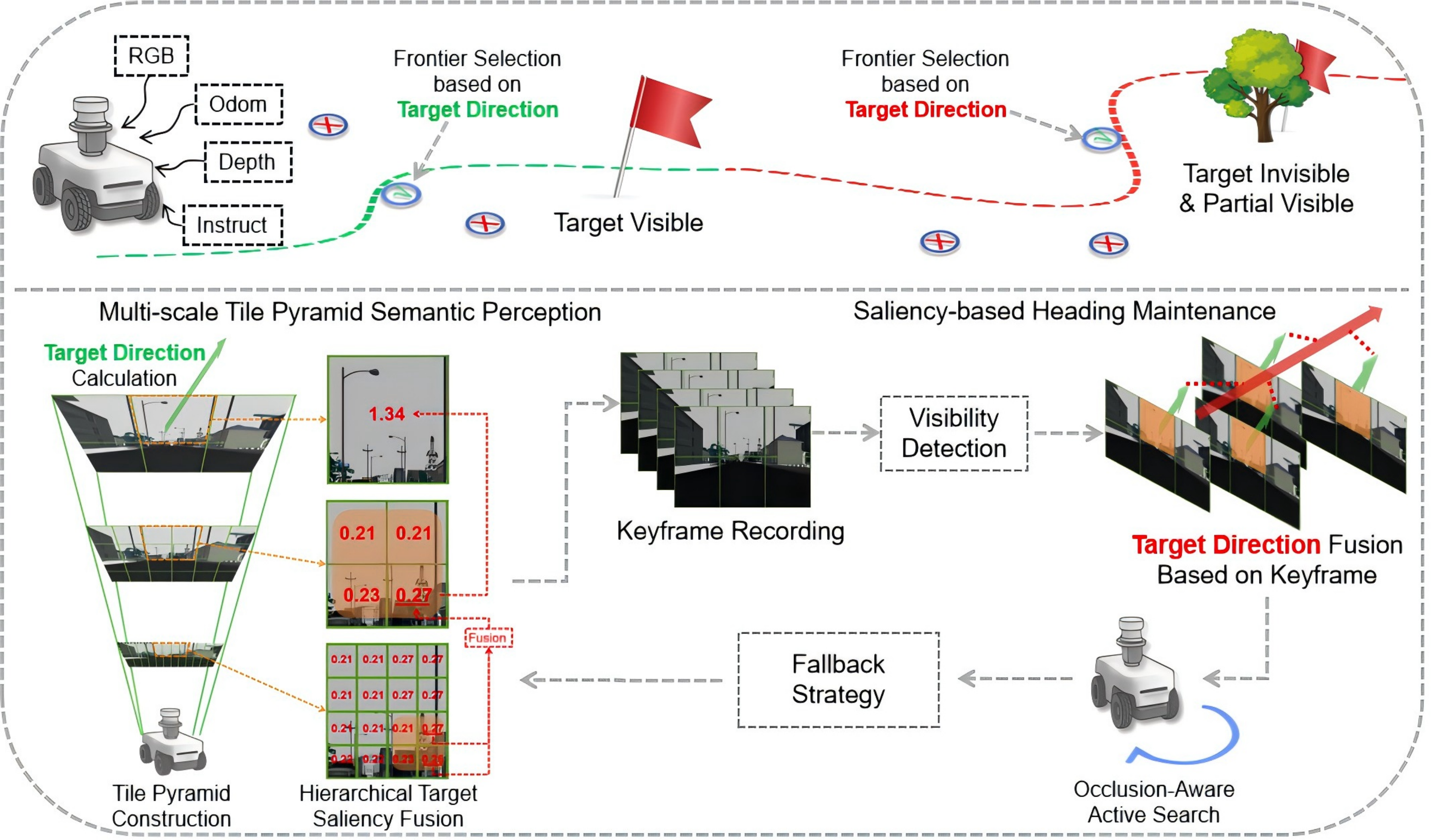}
    \caption{System overview. Multi-scale semantic perception and saliency-driven heading maintenance form a closed loop for outdoor ZSON under dynamic visibility: RGB + instruction yields target direction and visibility, while depth + odometry provide an occupancy grid for obstacle avoidance and frontier detection. Direction cues guide frontier selection and keyframe recording when the target is visible, and activate direction fusion, active search, and fallback strategies when the target is partially or fully occluded.}
    
    \label{pipeline}
    \vspace{-0.75cm}
\end{figure*}

\subsection{System Overview}
\label{sec:sysoverview}
To tackle the unified challenge of outdoor ZSON—pursuing distant targets that appear as tiny projections while remaining robust to dynamic visibility changes—we design a lightweight perception–navigation system that integrates multi-scale semantic amplification with saliency-driven heading maintenance. 

As illustrated in Fig.~\ref{pipeline}, the system forms a closed perception–decision loop. The perception module analyzes RGB observations through a multi-scale tile pyramid, amplifying weak semantic cues from tiny, long-range targets and estimating both target direction and visibility status. These outputs are consumed by the navigation module, which adapts robot behavior accordingly: when the target is visible, the estimated direction directly guides frontier selection (frontiers are the free–unknown boundaries on the occupancy grid) and forward progress; when the target becomes partially visible or fully occluded, the system leverages stored keyframes to sustain orientation, while triggering active search or fallback strategies for reliable re-identification.

Together, they close the loop for outdoor ZSON: the aligned multi-scale tile hierarchy makes regional semantic saliency reliable for far-field perception, and the same saliency sustains heading under variant visibility over time.

\subsection{Multi-scale Tile Pyramid Semantic Perception}
\label{sec:tile_pyramid}
As the perception backbone of our system, the multi-scale tile pyramid departs from image-wide heatmap reasoning~\cite{liu2024grounding} and instead adopts a hierarchical region-level comparison strategy. By partitioning the input into multi-scale tiles and progressively accumulating semantic saliency across scales, the system narrows the search space and amplifies weak cues from tiny, distant targets.

\textbf{Tile Pyramid Construction.} 
Given an RGB image $I$, we construct a three-level spatial pyramid $\mathcal{T}=\{T_1,T_2,T_3\}$—an aligned multi-scale tile hierarchy over a single-resolution image (region-wise tiling rather than multi-resolution rescaling), corresponding to fine ($T_1$: $8\times12$), medium ($T_2$: $4\times6$), and coarse ($T_3$: $2\times3$) partitions. This configuration can be flexibly adjusted for different scenes. Each level $T_l$ contains a set of image tiles $t_l^{(i)}$, where $i$ indexes the tile at layer $l$. For each tile, we employ CLIP~\cite{radford2021learning} to encode $t_l^{(i)}$ and compute the semantic similarity $s_l^{(i)}$ with the target prompt. Importantly, the three levels are dyadically aligned: each tile in $T_2$ exactly covers a fixed $2{\times}2$ block of children in $T_1$, and each tile in $T_3$ covers a fixed $2{\times}2$ block in $T_2$. Hence any coarse tile deterministically corresponds to a $4{\times}4$ region (16 tiles) in $T_1$, making the upward attribution from fine to coarse unique and deterministic (see the bottom-left panel of Fig.~\ref{pipeline}). 

Unlike classic image pyramids~\cite{he2015spatial,lin2017feature} that rescale the whole image at multiple resolutions, our design keeps a single-resolution image and performs region-wise partition with aligned tiles. This aligned, single-resolution partition induces fixed sibling sets for each parent (e.g., a $4{\times}4$ set of $T_1$ tiles for every coarse tile), which are natural units for local contrast and statistics. Intuitively, when a tiny target falls inside a coarse tile’s $4{\times}4$ child set in $T_1$, one or two tiles attain much higher similarities $s_1^{(i)}$ than their siblings, yielding high intra-set variance; without the target, scores are near-uniform and the variance stays low. This localized semantic contrast is the evidence we rely on; the next section aggregates it across scales.

\textbf{Hierarchical Target Saliency Fusion.} Each parent tile at level $l{+}1$ aggregates a fixed $2{\times}2$ child set at level $l$ and propagates its localized semantic contrast bottom-up via a variance-weighted residual update with exponential amplification $\beta=\text{base}^{\hat{\sigma}}$ (e.g., \textit{base}$=1.5$), where $\sigma$ is the standard deviation of scores within the child set (clipped to $\hat{\sigma}\in[0,1]$).

\textit{$T_1 \!\rightarrow\! T_2$ (fine $\rightarrow$ medium).}
For each parent tile in $T_2$, we take its $2{\times}2$ child tiles in $T_1$, select the two highest scores (Top-2), and update:
\begin{equation}
s_2' \;=\; s_2 \;+\; \beta \cdot \sum\nolimits_{\text{Top-2}} \, s_1 
\end{equation}

Here, Top-2 helps preserve small distant targets that may span multiple adjacent fine-scale tiles.

\textit{$T_2 \!\rightarrow\! T_3$ (medium $\rightarrow$ coarse).}
Similarly, for each parent tile in $T_3$, we take its $2{\times}2$ children in $T_2$, compute $\hat{\sigma}$ as above, and use Top-1 aggregation:
\begin{equation}
s_3' \;=\; s_3 \;+\; \beta \cdot \sum\nolimits_{\text{Top-1}} \, s_2' 
\end{equation}
At this stage, medium-scale tiles typically already cover the whole target, so Top-1 is more discriminative.

Variance therefore acts as a confidence modulator: groups with clear semantic contrast are amplified (via $\text{base}^{\hat{\sigma}}$), flat groups are down-weighted, yielding a stable coarse-layer \emph{regional saliency} that directly supports target direction estimation and visibility diagnosis.

\textbf{Target Direction Calculation.}
Using the coarse-level maximum in $T_3$ as the anchor, we take its center pixel $\mathbf{p}_\text{img}$ and, with intrinsics $K$ and extrinsics $T_{c\to w}$, obtain the world-frame target direction:
\begin{equation}
\hat{\mathbf{d}}_\text{target} \;=\; T_{c \to w}\, K^{-1}\, \tilde{\mathbf{p}}_\text{img}
\end{equation}

where $\tilde{\mathbf{p}}_\text{img}$ is the homogeneous form of $\mathbf{p}_\text{img}$. Here $T_{c\to w}$ is given by the calibrated extrinsics and the current odometry. The vector $\hat{\mathbf{d}}_\text{target}$ is then used for navigation heading maintenance during visibility change. This coarse-level anchor also provides the saliency evidence used for visibility diagnosis (Sec.~\ref{sec:visibility}), keeping direction and visibility consistent by design.

\subsection{Saliency-based Heading Maintenance}
\label{sec:visibility}
Building on the outputs of the perception backbone, the saliency-based heading maintenance module enables the system to sustain orientation and progress when target visibility becomes unstable. It records keyframes during visible observations and leverages heading memory, fusion, and active search to maintain continuity under partial or complete occlusion.

\textbf{Keyframe Recording.}  
To enhance navigation robustness when the target undergoes varying visibility conditions, we introduce a saliency-driven keyframe recording and heading maintenance mechanism. When the target is clearly visible and a global heading is inferred via semantic scoring and geometric projection, the current observation is stored as a keyframe $K_i$. Each keyframe contains:  
\begin{enumerate}
    \item Semantic Saliency Features: The most salient tile region in the image, along with its similarity score $s_t$, local mean $\mu_t$, and standard deviation $\sigma_t$, serving as a compact semantic descriptor of the target’s presence.
    \item Global Target Direction $\hat{\mathbf{d}}_\text{target}$: The center of the highest-scoring tile is projected into the world coordinate frame to obtain a normalized 3D vector pointing toward the target.
    \item Robot Observation Pose: The robot’s 6-DoF pose in the world frame at the time of observation, for later use in directional fusion and backtracking.
\end{enumerate}

All keyframes are maintained in a sliding window:
\begin{equation}
\{K_{i-N+1}, \dots, K_i\}
\end{equation}
which is used for directional fusion, frontier selection guidance, and semantic re-identification under occlusion.

\textbf{Visibility Detection and Direction Fusion.}  
We determine target visibility by analyzing the semantic score distribution across multiple levels of the tile pyramid. For each level $l$, we compute the mean $\mu_{s_l^{(i)}}$, standard deviation $\sigma_{s_l^{(i)}}$, and sparsity ratio:
\begin{equation}
r_l = \frac{\max(s_l^{(i)})}{\mu_{s_l} + \epsilon}
\end{equation}
where $s_l^{(i)}$ is the score of the $i$-th tile at level $l$, and $\epsilon$ is a small constant to prevent division by zero.  
If there exists a level satisfying $r_l > w_{r_l}  \land  \sigma_l > w_{\sigma_l}$ ($w_{r_l}$ and $w_{\sigma_l}$ are the sparsity ratio and standard deviation thresholds for target visibility, adopted from~\cite{zhang2025apexnav} for their proven balance between sensitivity to saliency and robustness to noise.), we consider the target visible, indicating a prominent semantic peak (high $r_l$) and spatially sparse distribution (high $\sigma_l$). Otherwise, the target is deemed occluded.

When occlusion occurs, the robot refers to the historical heading vectors $\{\hat{\mathbf{d}}^{(j)}_{\text{target}}\}$ in the sliding window, fusing them via weighted averaging to obtain a smoothed heading $\bar{\mathbf{d}}_{\text{target}}$. We use saliency-derived, time-decayed weights (based on the recorded sparsity ratio $r$ and variance $\sigma$ in keyframes) and normalize them to sum to one, which prevents drift and reflects confidence over time. This allows the robot to continue exploring plausible frontier regions along a consistent heading even when the target is not visible.

\textbf{Visibility-Aware Active Search.}
When the target is deemed invisible, the robot navigates toward the frontier region corresponding to the most recent global heading recorded in keyframes. Upon arrival, it performs local scanning within a limited angular range (e.g., $\pm30^\circ$) while reconstructing a multi-scale tile pyramid to compute semantic scores. High-scoring tiles are compared against salient regions stored in historical keyframes, and a tile is recognized as the target only if it exhibits both high semantic similarity to the text prompt and strong feature-level consistency with previous observations. This integrated process ensures that active search and semantic verification are carried out jointly, enabling stable and accurate target rediscovery. 

\textbf{Fallback Strategy for Persistent Occlusion.}  
If the target cannot be re-identified after visiting multiple frontiers and performing active searches, the robot returns to the most recent keyframe’s position and performs a full 360° look-around to regain target visibility and resume navigation.

This strategy emulates human behavior when a target disappears: maintaining forward progress based on heading memory while actively shifting the viewpoint to search for the target. By integrating keyframe memory, visibility detection, direction fusion, and fallback search within a single lightweight module, this component complements the perception backbone and completes the closed perception–navigation loop.

\section{Experiment}
We evaluate the unified challenge with two complementary sets of experiments: (i) long-range semantic perception (can directional cues be extracted from tiny, distant targets beyond depth range?), and (ii) navigation under dynamic visibility (can orientation be maintained and targets recovered under partial/full occlusion?). Together they validate robust far-field perception and stable navigation continuity under visibility changes.

\subsection{Long-range Semantic Perception}
\begin{figure}[!t]
    \centering
    \includegraphics[width=\linewidth]{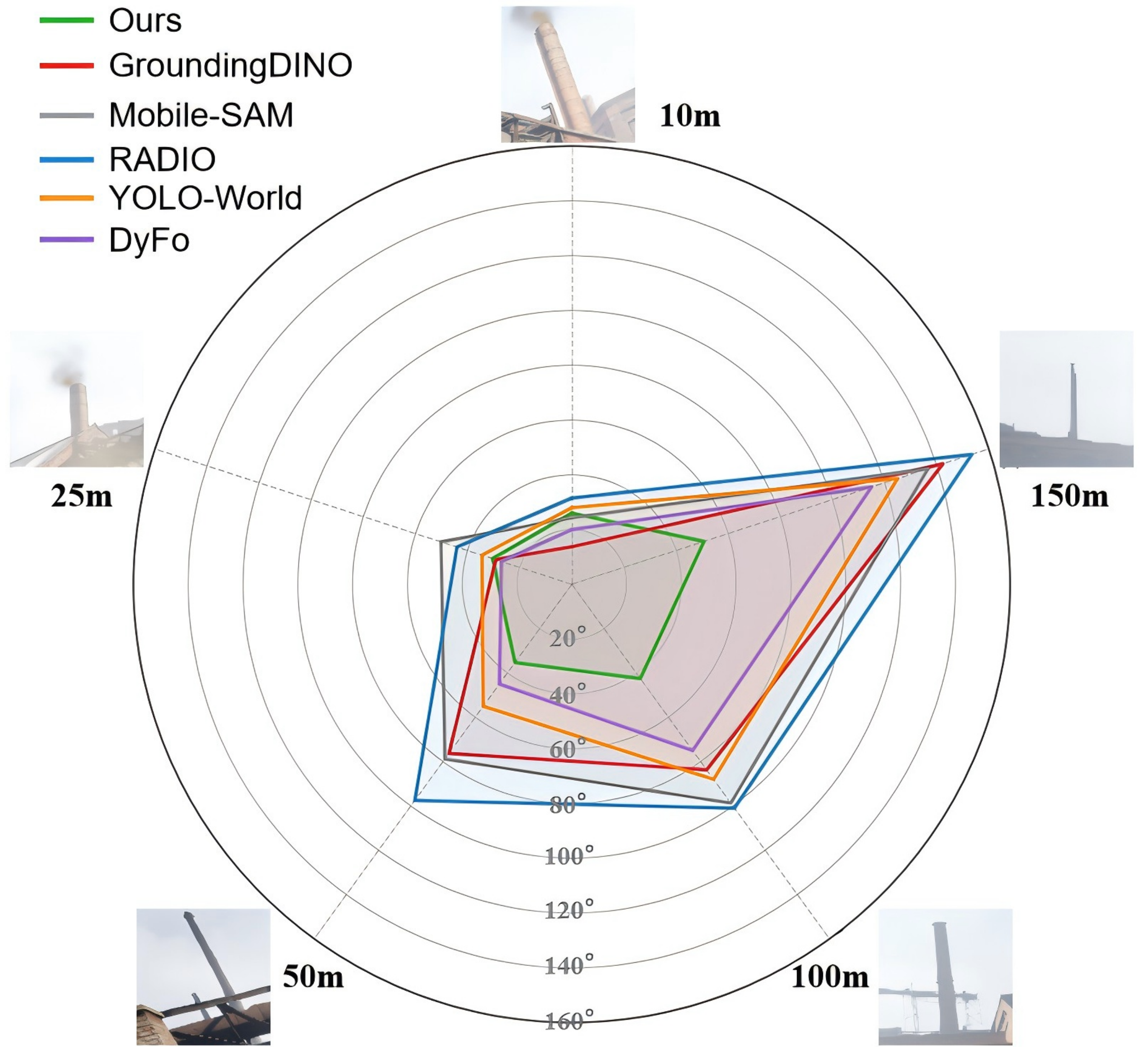}
    \caption{Comparison of penalized angular error across distances from 10\,m to 150\,m. Smaller values indicate better directional perception. Our method achieves lower error at longer ranges. Surrounding images visualize the increasing challenge from 10\,m to 150\,m.}
    \label{exp1}
        \vspace{-0.55cm}
            
\end{figure}

\begin{table}[t]
  \centering
      \vspace{0.6cm}
    \caption{Ablation study using penalized angular error $e_\text{avg}$. Ranking shown as \colorbox{green!25}{\textbf{first}}, \colorbox{yellow!30}{second}.}
  \label{tab:ablation_exp1}
  \resizebox{\columnwidth}{!}{
  \begin{tabular}{lccccc}
  \toprule
    Method Variant & 10m$\downarrow$ & 25m$\downarrow$ & 50m$\downarrow$ & 100m$\downarrow$ & 150m$\downarrow$ \\
    \midrule
    Full Method (Ours)              & \cellcolor{yellow!30}26.0 & \cellcolor{green!30}\textbf{30.5} & \cellcolor{green!30}\textbf{35.0} & \cellcolor{green!30}\textbf{42.5} & \cellcolor{green!30}\textbf{51.0} \\
    
    w/o Multi-scale Pyramid         & \cellcolor{green!30}\textbf{25.4} & 36.5 & 60.0 & 95.0 & 130.0 \\
    
    w/o Saliency Amplification      & 26.5 & \cellcolor{yellow!30}33.0 & 50.5 & 70.0 & 100.5 \\
    
    w/o Hierarchical Fusion         & 27.0 & 34.0 &\cellcolor{yellow!30} 42.0 & \cellcolor{yellow!30}55.0 & \cellcolor{yellow!30}70.5 \\
    \bottomrule
  \end{tabular}
  }
  \vspace{-0.75cm}
\end{table}
\begin{table*}[t]
    \centering
    \setlength{\tabcolsep}{5pt}  
    \caption{Navigation performance under different occlusion types. }
    \begin{tabular}{ccccccccccccc}
        \toprule
        \multirow{2}*{\textbf{Method}} & \multicolumn{3}{c}{\textbf{Short Invisibility}} & \multicolumn{3}{c}{\textbf{Long Invisibility}} & \multicolumn{3}{c}{\textbf{Mixed Visibility Changes}} \\
        \cmidrule(lr){2-4} \cmidrule(lr){5-7} \cmidrule(lr){8-10}
        & RSR $\uparrow$ & RPL $\downarrow$ & SR $\uparrow$ & RSR $\uparrow$ & RPL $\downarrow$ & SR $\uparrow$ & RSR $\uparrow$ & RPL $\downarrow$ & SR $\uparrow$ \\
        \midrule
        Fixed-Heading & 40.3 & \cellcolor{green!25}\textbf{4.5} & 5.5   &13.3   & 15.3  & 2.0  & 26.6  & 21.6  & 3.5      \\
        VLFM          & 55.4 & \cellcolor{yellow!30}5.2 & 8.0  &35.0   & \cellcolor{yellow!30}16.1  & 6.0 & 35.4  & 20.3  & 6.5     \\
        RayFront      & \cellcolor{yellow!30}75.5 & 5.6 & \cellcolor{yellow!30}27.0  &41.7   & \cellcolor{green!25}\textbf{14.4}  & \cellcolor{yellow!30}22.5 & \cellcolor{yellow!30}59.7  & \cellcolor{green!30}\textbf{15.9}  & \cellcolor{yellow!30}23.0     \\
        NoMaD         & 56.9 & 8.3 & 18.5  &\cellcolor{yellow!30}43.7   & 21.8  & 15.5 & 48.3  & 25.2  & 16.0     \\
        \textbf{Ours}& \cellcolor{green!25}\textbf{90.5} & 10.1  & \cellcolor{green!25}\textbf{42.0} 
                      & \cellcolor{green!25}\textbf{73.3} & 21.2  & \cellcolor{green!25}\textbf{38.5} 
                      & \cellcolor{green!25}\textbf{84.0} & \cellcolor{yellow!30}18.8 & \cellcolor{green!25}\textbf{40.5}\\
        \bottomrule
    \end{tabular}
    \vspace{-0.55cm}
    \label{occlusion_exp}

\end{table*}
\begin{table}[t]
  \centering
      \vspace{0.6cm}
  \caption{Ablation study on the navigation module under varying visibility conditions.}
  \label{tab:ablation_nav}
  \resizebox{\columnwidth}{!}{
  \begin{tabular}{lccc}
    \toprule
    Method Variant & RSR$\uparrow$ & RPL$\downarrow$ & SR$\uparrow$\\
    \midrule
    Full Method (Ours)              & \cellcolor{green!30}\textbf{82.6} & \cellcolor{yellow!30}22.7 & \cellcolor{green!30}\textbf{40.3} \\
    w/o Visibility Detection        & 52.4 & 41.2 & 12.7 \\
    w/o Direction Fusion            & \cellcolor{yellow!30}67.3 & 31.5 & 21.8 \\
    w/o Active Search               & 59.1 & \cellcolor{green!30}\textbf{18.4} & \cellcolor{yellow!30}23.6 \\
    \bottomrule
  \end{tabular}
  }
  \vspace{-0.55cm}
\end{table}
\subsubsection{Simulation Experimental Setup}
We evaluate long-range semantic perception using the Tartanair~\cite{wang2020tartanair} dataset, which provides images of targets captured at 10, 25, 50, 100, and 150 meters under varying outdoor visibility conditions. Eight targets are used, each annotated with ground-truth depth and a directional label.

\subsubsection{Baseline}
We compare against five representative and commonly used vision-language methods in object navigation tasks: (1) GroundingDINO~\cite{liu2024grounding}, a box-based vision-language model designed for open-vocabulary detection; (2) Mobile-SAM~\cite{zhang2023faster}, a lightweight promptable segmentation model; (3) RADIO~\cite{ranzinger2024radio}, which integrates features from CLIP, DINO, and SAM to support zero-shot semantic understanding; (4) YOLO-World~\cite{Cheng2024YOLOWorld}, an open-vocabulary extension of YOLOv8 that leverages CLIP text embeddings to enable prompt-driven detection with strong efficiency and small-object sensitivity; (5) DyFo~\cite{li2025dyfo}, a recently proposed dynamic focus framework that collaborates between LMMs and visual experts improve tiny object visual grounding. Although DyFo is originally designed for visual QA rather than detection, we adapt it by extracting its predicted focus regions and computing target directions from the region centers, enabling a fair comparison under our angular error metric.

\subsubsection{Evaluation Metrics}
To jointly evaluate the success rate and directional accuracy of each method, we adopt a penalized angular error metric $e_{\text{avg}}$, defined as:

\begin{equation}
e_{\text{avg}} = \frac{1}{N} \sum_{i=1}^{N}
\begin{cases}
|\hat{\theta}_i - \theta_i|, & \text{if detected} \\
\pi, & \text{otherwise}
\end{cases}
\label{eq:angle_error}
\end{equation}

Here, $\hat{\theta}_i$ and $\theta_i$ denote the predicted and ground-truth target direction in radians for the $i$-th image, respectively. If the target is not successfully detected, a fixed penalty of $\pi$ is assigned. This metric jointly evaluates both detection reliability and directional accuracy, the latter being crucial when targets lie beyond depth range.
All methods compute directional predictions using the same formulation as described in~\ref{sec:tile_pyramid}.

\subsubsection{Result}
As shown in Fig.~\ref{exp1}, all methods perform well at 10m–25m, where GroundingDINO and DyFo achieve the lowest errors. Beyond 50\,m, however, baseline performance drops sharply due to target shrinkage and unstable visibility, with frequent detection failures triggering the $\pi$ penalty. Box- and mask-based models (GroundingDINO, Mobile-SAM) quickly lose reliability; RADIO produces blurred heatmaps without explicit direction modeling; YOLO-World maintains moderate stability but suffers reduced recall; and DyFo’s focus tree fails to converge under tiny or occluded targets. In contrast, our method sustains low error up to 150\,m by leveraging multi-scale tiles and saliency scoring, clearly outperforming all baselines in long-range perception.

\subsubsection{Ablation Study}
To assess the contribution of each component in our framework, we further conduct ablation experiments by progressively disabling key modules. Results are summarized in Table~\ref{tab:ablation_exp1}. Removing the multi-scale pyramid leads to drastic degradation at long ranges (100\,m, 150\,m), indicating the necessity of hierarchical perception for tiny objects. Excluding saliency amplification causes moderate performance loss in the mid-to-far range, showing that variance-based boosting is crucial to highlight sparse cues. Without hierarchical fusion, errors increase particularly at 50\,m, as the lack of bottom-up information transfer weakens coarse-level confidence. These results confirm that each module plays a complementary role, and the full model achieves the most robust performance across distances.

\subsection{Navigation under Dynamic Visibility}
\label{sec:exp2}
To complement the perception benchmark, we further evaluate the system’s ability to sustain navigation when target visibility becomes unstable. While the previous experiments focused on extracting reliable long-range cues from tiny distant targets, this section examines how the system leverages these cues under dynamic visibility conditions.

\begin{figure*}[!t]
    \centering
    \includegraphics[width=\textwidth]{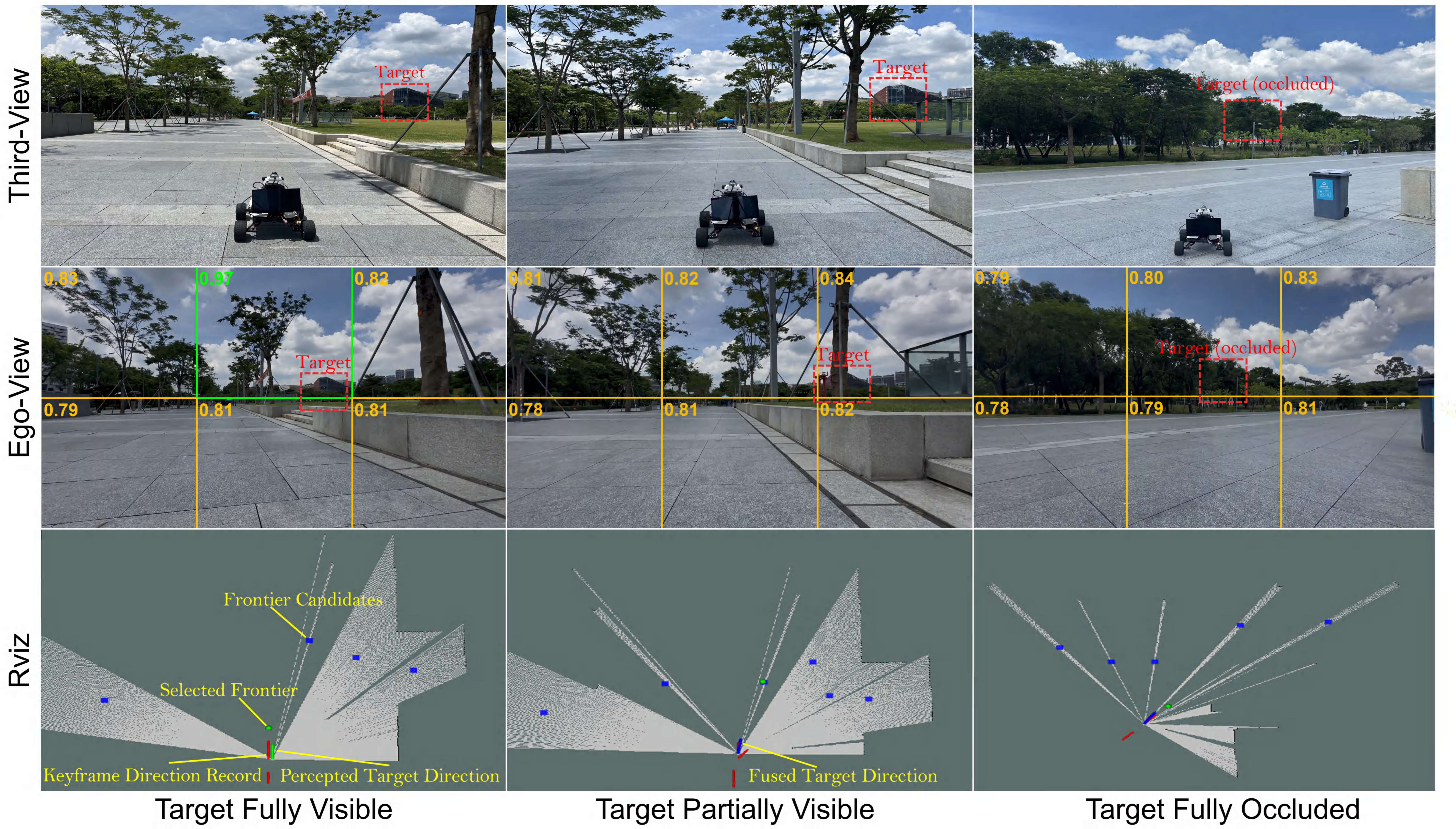}
    \caption{Visualization of real-world navigation under dynamic visibility.}

    \label{realworldexp1}
    \vspace{-0.55cm}
\end{figure*}

\begin{figure}[!t]
    \centering
    \includegraphics[width=\linewidth]{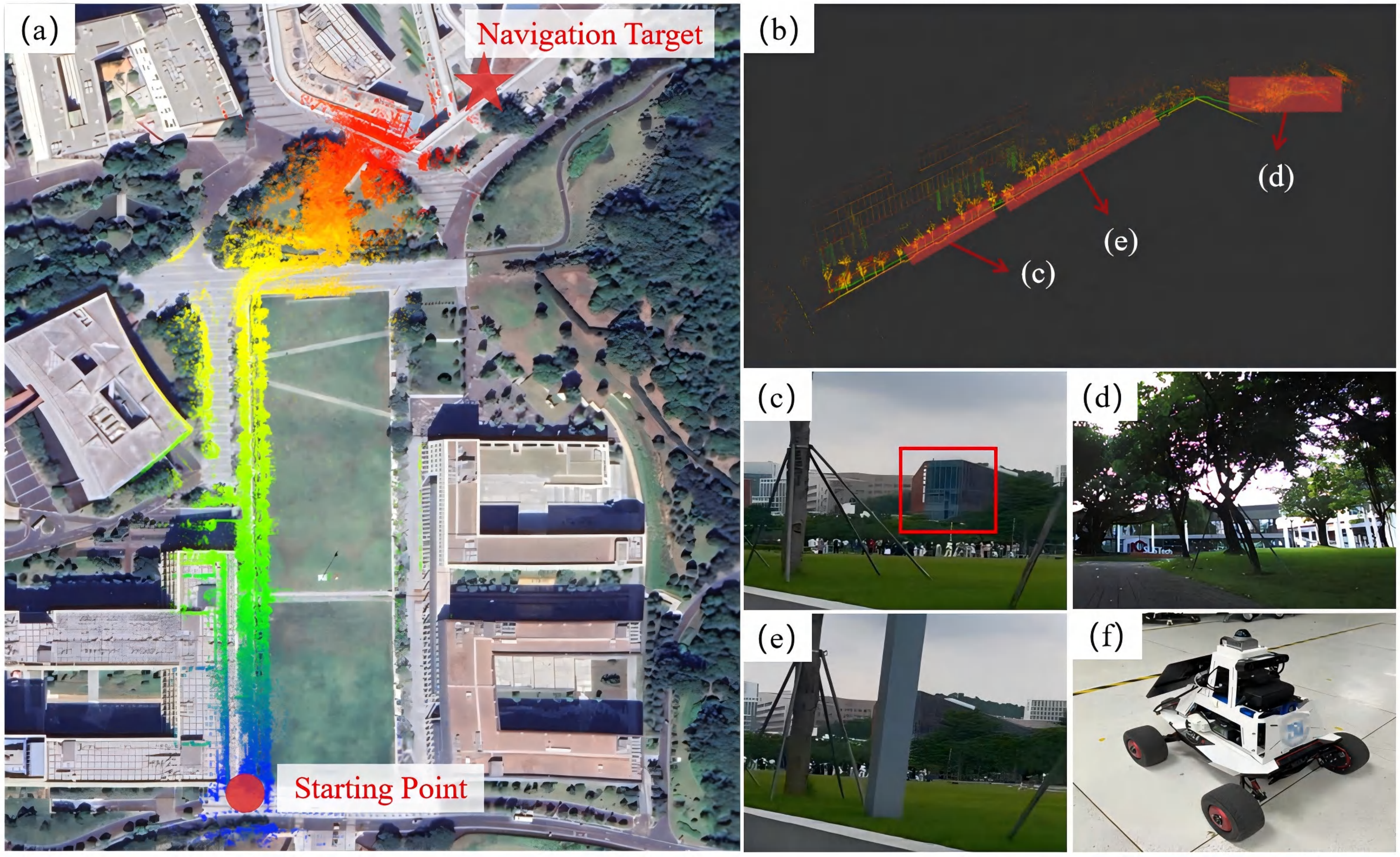}
    \caption{Real-world navigation experiment in a 350\,m\,$\times$\,450\,m campus environment.
(a) Navigation trajectory projected onto a satellite map, showing the robot’s path from the starting point (red circle) to the navigation target (red star), along with the projected LiDAR map.
(b) Mapping result during navigation, with red rectangles and arrows marking the robot’s observations along the route.
(c) Example of the target being fully visible, highlighted by a red bounding box.
(d) Example of the target being fully occluded.
(e) Example of the target being partially visible.
(f) The ground robot platform used for the experiment.}
    \label{realworldexp2}
    \vspace{-0.75cm}
\end{figure}
\subsubsection{Simulation Experimental Setup}
We use a 200\,m$\times$170\,m outdoor simulation with typical occluders (buildings, billboards, shrubs; Fig.~\ref{whatwedo}). Each episode starts $>$50\,m from the target. We design three occlusion types: (1) short ($\sim$3\,s), (2) long ($\sim$8\,s), and (3) mixed. The robot must maintain heading and actively recover target observations during visibility loss.

\subsubsection{Baseline} 
We compare against four methods: RayFront~\cite{alama2025rayfronts}, which projects long-range semantics for frontier exploration; NoMaD~\cite{sridhar2024nomad}, a diffusion-policy-based end-to-end navigation model; VLFM~\cite{yokoyama2024vlfm}, which fuses semantic heatmaps across frames (mainly for indoor tasks); and a Fixed-Heading baseline that simply maintains the last observed direction without recovery.

\subsubsection{Evaluation Metrics}  
We evaluate the performance using three key metrics. \textbf{Recovery Success Rate (RSR)} measures whether the robot can successfully re-observe the target and resume navigation after invisibility, reflecting heading maintenance ability. \textbf{Recovery Path Length (RPL)} denotes the accumulated path length between the moment the target disappears and when it is re-observed, indicating the efficiency of navigation under visibility loss. \textbf{Success Rate (SR)} reflects whether the robot ultimately reaches within 5 meters of the target, serving as an indicator of overall task completion.

\subsubsection{Result}
As shown in Table~\ref{occlusion_exp}, our method consistently achieves higher recovery success rates across all three visibility-change types. The improvement is especially notable in long and mixed occlusion scenarios, highlighting the robustness of our approach under sustained perceptual loss. Although our look-around behavior introduces some path redundancy, the recovery path length remains within a reasonable range, demonstrating a good trade-off between robustness and efficiency. It is worth noting that despite our method's superior occlusion recovery capability, the overall navigation success rate is still influenced by several factors, such as time constraints, target detection failures, or motion execution errors. These failure cases are further analyzed in the next section.

\subsubsection{Failure Case Analyze}
Although our method achieves consistently high RSR under various occlusion scenarios, it does not always guarantee a high overall navigation SR.
First, when occlusions persist for extended durations (e.g., over 10 seconds), heading memory can maintain a general orientation but active search often expands into overly large regions, resulting in delayed re-identification and trajectory overshoot.
Second, short but frequent occlusions create appearance ambiguities: the robot may pass near the target without recognition if the visible window is too brief to form a confident semantic match.
Third, repeated recovery cycles accumulate path length and increase the risk of exceeding time or step constraints, revealing a trade-off between robustness and efficiency.
Finally, occasional semantic misclassification or navigation-level errors (e.g., collision) further hinder task completion.
These failure cases indicate that while saliency-based heading maintenance ensures robustness in most occlusion scenarios, closing the gap between high RSR and high SR will require more advanced perception–planning integration, which we identify as an important direction for future work.

\subsubsection{Ablation Study}
We further ablate three core components of the navigation module: visibility detection, direction fusion, and active search. Without visibility detection, the system cannot reliably distinguish between partial occlusion and true absence, leading to frequent misjudgments and a drastic drop in recovery success. Removing direction fusion makes heading estimation unstable, causing larger path deviations and lower task success. Disabling active search prevents the robot from proactively re-observing the target, which reduces recovery success but shortens the recovery path due to the absence of scanning behavior. These results highlight the complementary roles of visibility-aware detection, heading smoothing, and proactive re-observation in achieving robust navigation across varying visibility states.

\subsection{Real-world Experiment}
To complement the simulation results and validate the system under real-world complexity, we conduct a large-scale outdoor navigation experiment in a 350\,m\,$\times$\,450\,m campus environment. This setting poses three major challenges aligned with our unified problem: (i) long-range navigation toward a distant goal, (ii) sparse semantic cues in open spaces, and (iii) frequent visibility changes caused by buildings, vegetation, and terrain. The robot is instructed to reach a distant library based on a language prompt. We use a Scout-Mini platform equipped with a stereo camera, a MID360 LiDAR, and an onboard Intel NUC 11 for real-time perception, mapping, and navigation.

To illustrate how the system handles different visibility states during navigation, Fig.~\ref{realworldexp1} provides qualitative results. The first row shows third-person views as the robot progresses toward the target, which transitions from fully visible to partially visible and finally fully occluded. The second row depicts the robot’s egocentric observations, where the multi-scale tile pyramid assigns semantic scores: green boxes indicate confident target perception, while orange boxes denote low-scoring background tiles. The third row visualizes the decision-making process in RViz. Frontier candidates are shown in blue, while arrows indicate directional guidance: green for currently perceived target direction, red for recorded keyframe direction, and blue for fused direction under occlusion. The selected frontier (green dot) aligns with these cues, demonstrating how the robot maintains orientation and exploration progress across visibility transitions.

As shown in Fig.~\ref{realworldexp2}, the robot successfully completes long-range navigation while the distant target undergoes multiple visibility changes. Throughout the trajectory, the system maintains reliable target perception and stable heading, confirming the effectiveness of our unified framework for outdoor ZSON in real-world deployment.

\section{Conclusion, Limitation and Future Work}
In this work, we presented a zero-shot framework for navigating toward distant prompted targets in large-scale outdoor environments with frequent occlusions. Our approach integrates a multi-scale tile pyramid for long-range semantic perception without relying on depth input, together with a saliency-based heading maintenance strategy for robust navigation under varying visibility. The system demonstrates reliable performance in both simulation and real-world settings.

Nevertheless, the framework has notable limitations. It still struggles under long-term occlusions and relies on an initial observation of the target to bootstrap navigation, making recovery difficult in complex outdoor scenarios. Moreover, the current path planning module lacks advanced reasoning capabilities, which limits the overall success rate when navigating through highly cluttered or uncertain environments.

To address these challenges, future work will explore integrating robotic reasoning mechanisms that fuse historical observations, semantic cues, and environmental context for more informed frontier selection and path planning.






\bibliographystyle{./bibliography/IEEEtran}
\bibliography{./bibliography/ref}
\clearpage

\end{document}